\documentclass[11pt,a4paper]{article}
\usepackage[hyperref]{emnlp-ijcnlp-2019}
\usepackage{times}
\usepackage{latexsym}

\usepackage{url}

\usepackage{footnote}
\usepackage{xcolor}
\usepackage{url}
\usepackage{graphicx}
\usepackage{booktabs}
\usepackage{multirow}
\usepackage{amsmath}
\usepackage{color, soul}
\aclfinalcopy

\title{A Summarization System for Scientific Documents}

\author{Shai Erera$^1$, Michal Shmueli-Scheuer$^1$, Guy Feigenblat$^1$, Ora Peled Nakash$^2$, \\
\textbf{Odellia Boni$^1$, Haggai Roitman$^1$, Doron Cohen$^1$, Bar Weiner$^1$, Yosi Mass$^1$, Or Rivlin$^1$},~\\
\textbf{Guy Lev$^1$, Achiya Jerbi$^1$, Jonathan Herzig$^1$, Yufang Hou$^1$, Charles Jochim$^1$},~\\
\textbf{Martin Gleize$^1$, Francesca Bonin$^1$, Debasis Ganguly$^1$, David Konopnicki$^1$}\\
\mbox{}\\
$^1$IBM Research, $^2$IBM Cloud \\
\small{\texttt{davidko@il.ibm.com}}
}

\date{}

\begin{document}
\maketitle
\begin{abstract}
We present a novel system providing summaries for Computer Science publications. 
Through a qualitative user study, we identified the most valuable scenarios for discovery, exploration and understanding of scientific documents.
Based on these findings, we built a system that retrieves and summarizes scientific documents for a given information need, either in form 
of a free-text query or by choosing categorized values such as scientific tasks, datasets and more.
Our system ingested 270,000 papers, and its summarization module aims to generate concise yet detailed summaries. We validated our approach with human experts.

\end{abstract}

\section{Introduction}
\label{intro}

The publication rate of scientific papers is ever increasing and many tools such as Google Scholar, Microsoft Academic and more, provide search capabilities and allow researchers to find papers of interest.
In Computer Science, and specifically, natural language processing, machine learning, and artificial intelligence, new tools that go beyond search capabilities are used to monitor\footnote{{\url{arxiv-sanity.com}}}, explore~\cite{CLscholar}, discuss and comment\footnote{\url{groundai.com/}} publications. 
Yet, there is still a high information load on researchers that seek to keep up-to-date.
Summarization of scientific papers can mitigate this issue and expose researchers with adequate amount of information in order to reduce the load.

Many tools for text summarization are available\footnote{\url{github.com/miso-belica/sumy,ivypanda.com/online-text-summarizer}}. However, such tools target mainly news or simple documents, not taking into account the characteristics of scientific papers i.e., their length and complexity. 

A summarization system for scientific publications requires many underlying technologies: first, extracting structure, tables and figures from PDF documents, then, identifying important entities, and, finally,  generating a useful summary.
We chose to provide summarization as part of a search system as it is the most common interface to consume scientific content, regardless of the task. 

\paragraph{Use-cases.}
\label{usecases}
We identified the most valuable scenarios for scientific paper usage through a qualitative user study.
We interviewed six potential users: a PhD student, two young researchers, two senior researchers, and a research strategist, all in the NLP domain. Users were asked to describe when do they access scientific papers, how often does it happens, how do they explore content, and finally, what are their pain-points with current tools. Top scenarios were, by order of frequency, (1) keeping updated on current work, (2) preparing a research project/grant request, (3) preparing related works when writing a paper, (4) checking the novelty of an idea, and (5) learning a new domain or technology. While (2), (3), (4), and (5) are important, it seems that they happen only a few times a year, whereas scenario (1) happens on a daily/weekly basis. All users mentioned information overload as their main problem, and, foremost, the efforts incurred by reading papers. Thus, we decided to focus on scenario (1).  
We further asked the users to describe: (a) how do they search and (b) the strategy they use to decide whether they want to read a paper.
For (a), users mentioned searching by using either keywords, entities (e.g., task name, dataset name, benchmark name), or citation. In this scenario, users are familiar with their research topic, and hence can be very focused. Some examples queries were ``state of the art results for \textit{SQUAD}'' or ``using \textit{BERT} in \textit{abstractive summarization}''.
For (b), users first read the title, and if relevant, continue to the abstract. Here, users mentioned, that in many cases, they find the abstract not informative enough in order to determine relevance. Hence the importance of summarization for helping researchers understand the gist of a paper without the need to read it entirely or even opening the PDF file.

\paragraph{Approach and Contribution.}

We present a novel summarization system for Computer Science publications, named \textsf{IBM Science Summarizer}, which can be useful foremost to the ACL community, and to researchers at large. It produces summaries focused around an information need provided by the user - a natural language query, scientific tasks (e.g., ``Machine Translation''), datasets or academic venues. \textsf{IBM Science Summarizer} summarizes the various sections of a paper independently, allowing users to focus on the relevant sections for the task at hand. In doing so, the system exploits the various entities and the user's interactions, like the user query, in order to provide a relevant summary. We validated our approach with human experts. The system  is available at: 
\url{https://ibm.biz/sciencesum}.

\section{Related Work}
\label{related}
Numerous tools support the domain of scientific publications including search, monitoring, exploring and more.
For automatic summarization, efforts mostly concentrated on automated generation of survey papers~\cite{Surveyor,CitationAS}. Surveyor~\cite{Surveyor} considers both content and discourse of source papers when generating survey papers. CitationAS~\cite{CitationAS} automatically generates survey papers using citation content for the medical domain.
The main differences between these systems and ours is that they create summaries from multi-documents, while our tool summarizes individual papers and supports query-focused summaries.

For supporting the ACL community, CL Scholar~\cite{CLscholar} presents a graph mining tool on top of the ACL anthology and enables exploration of research progress. TutorialBank~\cite{TutorialBank} helps researchers to learn or stay up-to-date in the NLP field.
Recently, paperswithcode\footnote{\url{paperswithcode.com/}} is an open resource for ML papers, code and leaderboards. 
Our work is complementary to these approaches and provide the first tool for automatic summarization and exploration of scientific documents.\footnote{For clarity, more related works are referred to in the various sections of this paper.}

\section{System Overview}
\label{framework}
\begin{figure}[t]
\begin{center}
  \includegraphics[width=0.55\columnwidth]{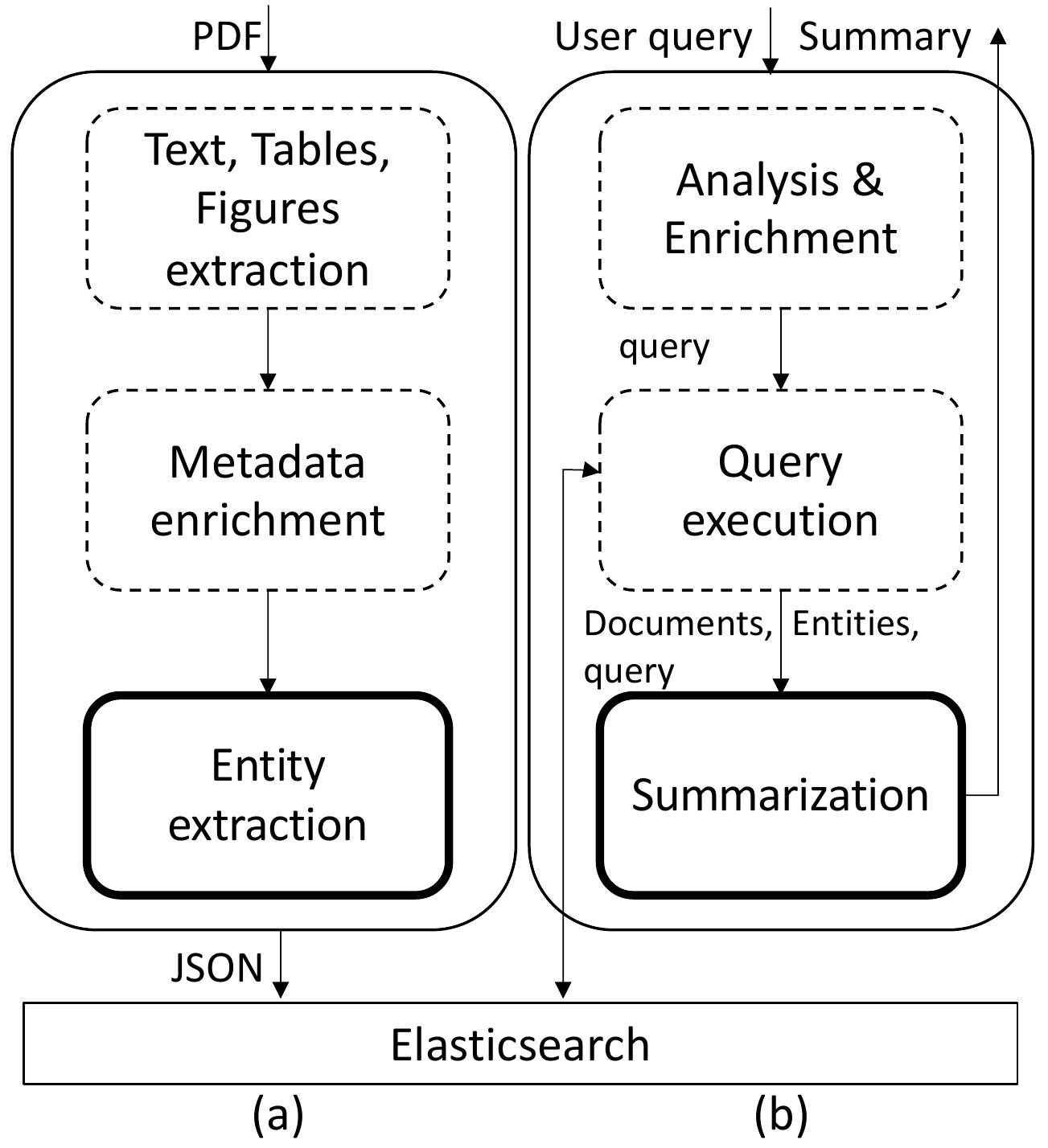}
 \end{center}
  \caption{\textsf{IBM Science Summarizer} Framework.}~\label{fig:framework}
\end{figure}

\textsf{IBM Science Summarizer}'s main purpose is to support discovery, exploration and understanding of scientific papers by providing summaries. 
The system has two parts. First, an ingestion pipeline parses and indexes papers' content from arXiv.com and ACL anthology, as depicted in Figure~\ref{fig:framework}(a). 
Second, a search engine (backed up by a UI), supports search and exploration, coupled with summarization, as shown in Figure~\ref{fig:framework}(b).

\begin{figure*}
\begin{center}
\includegraphics[width=0.80\textwidth]{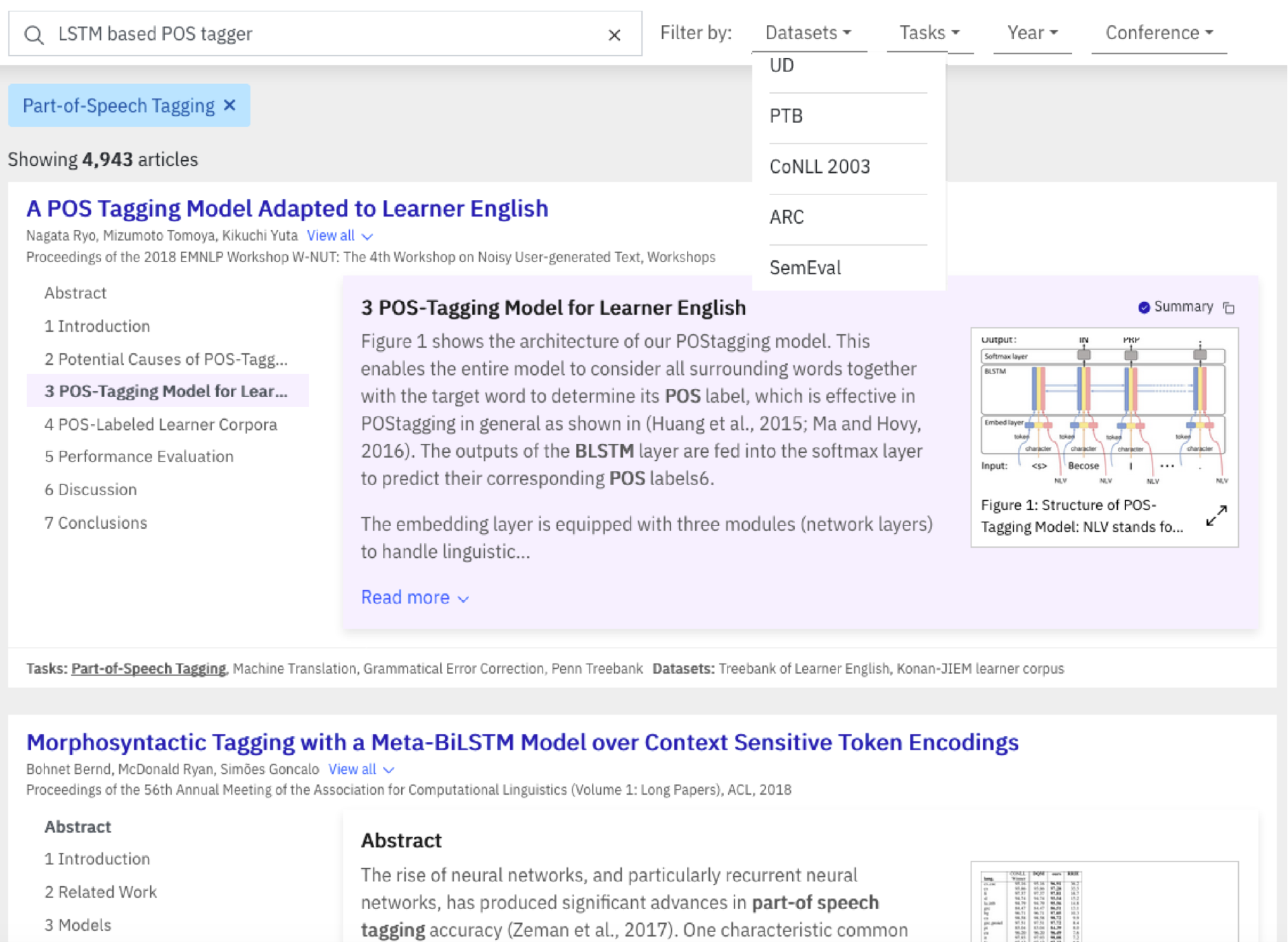}
\end{center}
\caption{\textsf{IBM Science Summarizer} UI.}~\label{fig:ui}
\end{figure*}

Figure~\ref{fig:ui} shows the user-interface for \textsf{IBM Science Summarizer}. Users interact with the system by posing natural language queries, or by using filters on the  metadata fields such as conference venue, year, and author, or entities (e.g., tasks, datasets)\footnote{In this case, there is no user query.}. User experience is an important usability factor. Thus, our UI provides indicators to help users explore and understand results. Specifically, associating a comprehensive structure with each result allows users to navigate inside content in a controlled manner: each section shows clearly the elements that are computed by the system (section summary, detected entities, etc.) and the elements that are directly extracted from the original paper. This clear distinction allows users to have visibility into the systems' contributions~\cite{AI_UI}.

\section{Ingestion Pipeline}
\label{indexing}
Our system contains 270,000 papers from arXiv.org (``Computer Science'' subset) and the ACL anthology\footnote{We removed duplication between the two by using Jaccard similarity on the titles and authors.}. The ingestion pipeline consists of paper acquisition, extracting the paper's text, tables and figures and enriching the paper's data with various annotations and entities.

\paragraph{Paper Parsing.}
We use Science-Parse\footnote{\url{github.com/allenai/science-parse}} to extract the PDF text, tables and figures. Science-Parse outputs a JSON record for each PDF, which among other fields, contains the title, abstract text, metadata (such as authors and year), and a list of the sections of the paper, where each record holds the section's title and text. We have merged sub-sections into their containing sections and this resulted in about 6-7 merged sections per article (e.g., see Fig.~\ref{fig:ui}). 
Science-Parse also supports extracting figures and tables into an image file, as well as caption text.

In addition, we detect figure and table references in the extracted text. We extract tasks, datasets and metric  (see details below).
Finally, we use Elasticsearch\footnote{\url{https://www.elastic.co}} to index the papers, where for each paper we index its title, abstract text, sections text and some metadata. 

\paragraph{Entities Extraction.}
\label{entities}
Entities in our system are of three types, \textit{task} (e.g., ``Question Answering''), \textit{dataset} (e.g., ``SQuAD2.0''), and \textit{metric} (e.g., ``F1''). We utilize both a dictionary-based approach and learning-based approach as follows.
First, we adopted the manual curated dictionaries of paperswithcode$^5$. 
Since those dictionaries may not cover all evolving topics, we further developed a module that automatically extracts entities.
Differently from previous work on information extraction from scientific literature which mainly focused on the abstract section \cite{semeval2018,Luan2018}, we analyze the entire paper and extract the above three types of entities that are related to the paper's main research findings. We cast this problem as a textual entailment task: we treat paper contents as \emph{text} and the targeting \emph{Task-Dataset-Metric (TDM)} triples as \emph{hypothesis}. The textual entailment approach forces our model to focus on learning the similarity patterns between \emph{text} and various triples. 
We trained our module on a dataset consisting of 332 papers in the NLP domain, and it achieves a macro-F1 score of 56.6 and a micro-F1 score of 66.0 for predicting TDM triples on a testing dataset containing 162 papers~\cite{TDM-acl2019}.
In total, our system indexed 872 tasks, 345 datasets, and 62 metrics from the entire corpus.

\section{Summarization}
\label{summarization}
This module generates a concise, coherent, informative summary for a given scientific paper that covers the main content conveyed in the text. The summary can either be focused around a query, or query agnostic (a generic summary)\footnote{Note that in order to optimize the response time, the production system currently offers query agnostic summaries.}. 
Scientific papers are complex: they are long, structured, cover various subjects and the language may be quite different between sections, e.g., the introduction is quite different than the experiments section. To ensure our summarizer assigns sufficient attention to each of these aspects we have opted to generate a standalone summary for each section. This way we summarize a shorter, more focused text, and the users can navigate more easily as they are given the structure of the paper. Each of these section-based summaries are eventually composed together into one paper summary.  

Scientific papers summarization goes back more than thirty years. Some of these works focus on summarizing content~\cite{Paice:1980:AGL:636669.636680, Paice:1993:IIC:160688.160696}, while others focused on citation sentences (citation-aware summarization) \cite{Elkiss:2008:BME:1331122.1331127, Qazvinian:2008:SPS:1599081.1599168,Abu-Jbara:2011:CCS:2002472.2002536}. Recently, \newcite{Yasunaga2019ScisummNetAL} released a large-scale dataset, \textit{Scisumm-Net},  including summaries produced by humans for over $1000$ scientific papers using solely the papers abstract and citations. While citations data encompasses the impact of the paper and views from the research community, it is not available for newly-published papers, and tends to lead to high level and shorter summaries (\textit{Scisumm-Net} average summary length is $151$ words). We opted to focus on more extensive, detailed summaries which do not rely on citations data. 
As mentioned above, the inputs to the summarization module are an (optional) query and entities (task, dataset, metric), and the relevant papers returned by the search/filtering (see Fig.~\ref{fig:ui}).
Given a retrieved paper and the optional query $Q$ (or entity), we describe next how a summary is produced for each section $D$ in the paper.

\paragraph{Query Handling.}
If present, $Q$ can either be short and focused or verbose. If short, it is expanded using query expansion~\cite{Xu:2009:QDP:1571941.1571954}. This pseudo-relevance feedback transforms $Q$ into a profile of $100$ unigram terms, obtained from analyzing the top papers that are returned from our corpus as a response to the given query. Alternatively, in the case of a verbose query, a \textit{Fixed-Point} term weighting schema~\cite{Paik:2014:FMW:2661829.2661957} is applied in order to rank the terms of the query. 

Alternatively, if only filtering is applied and there is no query, the keyphrases of the paper are extracted and used as a surrogate for the query. In this case, all keywords in the generated query are given the same weight.

\paragraph{Pre-Processing.}\label{sec:pre_processing}
Sentences are segmented using the \textit{NLTK} library and each sentence is tokenized, lower cased and stop words are removed. Then, each sentence is transformed into a unigrams and bi-grams \textit{bag-of-words} representations, where each \textit{n-gram} is associated with its relative frequency in the text.

\paragraph{Summarization Algorithm.}\label{sec:summarization}
In general, summaries can either be extractive or an abstractive. In the extractive case, a summary is generated by selecting a subset of sentences from the original input. Abstractive summarizers, on the other hand, can also paraphrase input text. In many cases, extractive summarization generates grammatical and focused summaries while abstractive techniques require heavy supervision, are limited to short documents and may transform meaning~\cite{Gambhir:2017:RAT:3041102.3041126}. 

In our system, summarization is applied on $D$ using a state-of-the-art unsupervised, extractive, query focused summarization algorithm, inspired by \cite{DBLP:conf/sigir/FeigenblatRBK17}, whose details are briefly described as follows.
The algorithm gets a paper section, a natural language query $Q$, a desired summary length (in our case, $10$ sentences\footnote{We leave the study of variable-length section summaries for future work.}), and a set of entities associated with the query $E_Q$. The output $S$ is a subset of sentences from $D$ selected through an unsupervised optimization scheme. To this end, the sentence subset selection problem is posed as a multi-criteria optimization problem, where several summary quality objectives are be considered. The selection is obtained using the \textit{Cross Entropy (CE)} method \cite{Rubinstein:2004:CEM:1014902}.  Optimization starts by assigning a uniform importance probability to each sentence in $D$. Then, CE works iteratively, and, at each iteration, it samples summaries using a learnt distribution over the sentences, and evaluates the quality of these summaries by applying a target function. This function takes into account several quality prediction objectives, which (for simplicity) are multiplied together. The learning process employs an exploration-exploitation trade-off in which the importance of a sentence is a fusion between its importance in previous iterations and its importance in the current one.

The following five summary quality predictors are used by~\newcite{DBLP:conf/sigir/FeigenblatRBK17}: query saliency, entities coverage, diversity, text coverage and sentence length.
\textit{Query saliency} measures to what extent the summary contains query related terms as expressed by the cosine similarity between the unigrams \textit{bag-of-words} representation of the summary and the query terms. \textit{Entities coverage} measures to what extent the set of entities identified in a summary shares the same set of entities with $E_Q$, measured by the Jaccard similarity between the sets. The aim of this objective is to produce a summary that is more aligned with the information need provided explicitly (as a filter specified by the user) or implicitly (learnt from the query terms). \textit{Diversity} lays towards summaries with a diverse language model using the entropy of the unigrams \textit{bag-of-words} representation of the summary. \textit{Text coverage} measures the summary coverage of $D$ as measured by cosine similarity between the bi-gram bag-of-words representation of a summary and $D$. Finally, the {length} objective biases towards summaries that include longer sentences, which tend to be more informative.

\section{Human Evaluation}
\label{eval}

\textsf{IBM Science Summarizer} summarization paradigm is section-based, i.e., each section is summarized independently, and then all sections' summaries are combined into the paper's summary. In order to evaluate this paradigm, we approached $12$ authors from the NLP community, and asked them to evaluate summaries of two papers that they have co-authored (preferably as the first author).
For each paper, we generated two summaries of two types: the \textit{section-based} summary as described above, and a second summary generated using the same algorithm but ignoring sections (i.e., treating the paper content as flat text), a \textit{section-agnostic} summary. For the section-based summary, each section's summary length was fixed to $10$ sentences.
The length of the section-agnostic summary was defined as the length of the section-based summary. In total $24$ papers, and $48$ summaries were evaluated.
\paragraph{Tasks.} The authors evaluated summaries of each summary type, section-agnostic and section-based (in random order), by performing the following $3$ tasks per summary: (1) for each sentence in the summary, we asked them to indicate whether they would consider it as a part of a summary of their paper (i.e., precision oriented measure); (2) we asked them to evaluate how well each of the sections of the paper is covered in the summary (i.e., coverage/recall); and (3)  we asked them to globally evaluate the quality of the summary.
For tasks (2) and (3) we used a 1-5 scale, ranging from very bad to excellent, $3$ means good. 
\paragraph{Analysis.} The objective of the analysis is to find quantitative scores for each summary type to facilitate a comparison between them. For task (1), for each paper, we calculated the precision scores of the two summary types, and then computed the average score for each summary type across all papers.
For task (2), we calculated an average score for each paper and summary type by averaging over the sections scores. Then, we obtained the average of these scores for each summary type across all papers.
Finally, for task (3), we simply averaged the scores given by the authors to each summary type. To further quantify the evaluation, we analyzed how well each summary type did for each of the 3 tasks. For that we counted the number of times that each summary type scored better than the other, and then divided by the total number of papers, to obtain the ``\% wins''.

\paragraph{Results.} Table~\ref{tab:results} summarizes the results across the 3 tasks. 
For example, for task (2), for $68\%$ of the papers, the section-based summary was scored higher, while, for $22\%$ the section-agnostic summary was scored higher (for $10\%$ of the papers, the summaries were scored equally). The average score for section-based summaries was $3.32$ with standard deviation of $0.53$.
Notably, the quality of the section-based summaries significantly outperforms the section-agnostic summaries on all 3 tasks, supporting our proposed paradigm.
\begin{table}[]
\centering
\resizebox{1.0\columnwidth}{!}{
\begin{tabular}{|c||c|c|c|c|}
\hline
Task        &  \multicolumn{2}{c|}{Section-agnostic}   &  \multicolumn{2}{c|}{Section-based}  \\\hline\hline
 &\% wins & Avg. score (std)&\% wins & Avg. score (std)   \\\hline
(1)                  &37 & 0.54 (0.17) &63&  0.6 (0.18)$\dagger$  \\ \hline
(2)                  & 22 &  3 (0.56) &68   &3.32 (0.53) $\dagger$  \\ \hline
(3)                  & 4.5 &  2.86 (0.56) &36& 3.22 (0.61) $\ddagger$  \\ \hline
\end{tabular}}
\caption{Tasks results for section-agnostic, and section-based. $\dagger$ - The results were significant with $p<0.05$. $\ddagger$- The results were significant with $p<0.005$. }
\label{tab:results}
\end{table}

\section{Conclusion}
\label{conclusion}

We presented \textsf{IBM Science Summarizer}, the first system that provides researchers a tool to systematically explore and consume summaries of scientific papers. 
As future work, we plan to add support for additional entities e.g., methods, and to increase our corpus to include more papers. Finally, we plan to provide this tool to the community as an open service and conduct an extensive user study about the usage and quality of the system, including automatic evaluation of the summaries.

\bibliography{references}

\begin{thebibliography}{19}
\expandafter\ifx\csname natexlab\endcsname\relax\def\natexlab#1{#1}\fi

\bibitem[{Abu-Jbara and Radev(2011)}]{Abu-Jbara:2011:CCS:2002472.2002536}
Amjad Abu-Jbara and Dragomir Radev. 2011.
\newblock \href {http://dl.acm.org/citation.cfm?id=2002472.2002536} {Coherent
  citation-based summarization of scientific papers}.
\newblock In \emph{Proceedings of the 49th Annual HLT}, HLT '11, pages
  500--509. Association for Computational Linguistics.

\bibitem[{Elkiss et~al.(2008)Elkiss, Shen, Fader, Erkan, States, and
  Radev}]{Elkiss:2008:BME:1331122.1331127}
Aaron Elkiss, Siwei Shen, Anthony Fader, G\"{u}ne\c{s} Erkan, David States, and
  Dragomir Radev. 2008.
\newblock \href {https://doi.org/10.1002/asi.v59:1} {Blind men and elephants:
  What do citation summaries tell us about a research article?}
\newblock \emph{J. Am. Soc. Inf. Sci. Technol.}, 59(1):51--62.

\bibitem[{Fabbri et~al.(2018)Fabbri, Li, Trairatvorakul, He, Ting, Tung,
  Westerfield, and Radev}]{TutorialBank}
Alexander Fabbri, Irene Li, Prawat Trairatvorakul, Yijiao He, Weitai Ting,
  Robert Tung, Caitlin Westerfield, and Dragomir Radev. 2018.
\newblock Tutorialbank: A manually-collected corpus for prerequisite chains,
  survey extraction and resource recommendation.
\newblock In \emph{Proceedings of the 56th ACL}, pages 611--620.

\bibitem[{Feigenblat et~al.(2017)Feigenblat, Roitman, Boni, and
  Konopnicki}]{DBLP:conf/sigir/FeigenblatRBK17}
Guy Feigenblat, Haggai Roitman, Odellia Boni, and David Konopnicki. 2017.
\newblock Unsupervised query-focused multi-document summarization using the
  cross entropy method.
\newblock In \emph{Proceedings of the 40th International {ACM} {SIGIR}}, pages
  961--964.

\bibitem[{Flavian et~al.(2009)Flavian, Gurrea, and Orus}]{AI_UI}
Carlos Flavian, Raquel Gurrea, and Carlos Orus. 2009.
\newblock Web design: a key factor for the website success.
\newblock \emph{Journal of Systems and Information Technology}, 11(2):168--184.

\bibitem[{G{\'{a}}bor et~al.(2018)G{\'{a}}bor, Buscaldi, Schumann, QasemiZadeh,
  Zargayouna, and Charnois}]{semeval2018}
Kata G{\'{a}}bor, Davide Buscaldi, Anne{-}Kathrin Schumann, Behrang
  QasemiZadeh, Ha{\"{\i}}fa Zargayouna, and Thierry Charnois. 2018.
\newblock Semeval-2018 task 7: Semantic relation extraction and classification
  in scientific papers.
\newblock In \emph{Proceedings SemEval@NAACL-HLT 2018}, pages 679--688.

\bibitem[{Gambhir and Gupta(2017)}]{Gambhir:2017:RAT:3041102.3041126}
Mahak Gambhir and Vishal Gupta. 2017.
\newblock \href {https://doi.org/10.1007/s10462-016-9475-9} {Recent automatic
  text summarization techniques: A survey}.
\newblock \emph{Artif. Intell. Rev.}, 47(1):1--66.

\bibitem[{Hou et~al.(2019)Hou, Jochim, Gleize, Bonin, and
  Ganguly}]{TDM-acl2019}
Yufang Hou, Charles Jochim, Martin Gleize, Francesca Bonin, and Debasis
  Ganguly. 2019.
\newblock \href {https://arxiv.org/abs/1906.09317} {Identification of tasks,
  datasets, evaluation metrics, and numeric scores for scientific leaderboards
  construction}.
\newblock volume arXiv:1906.09317.

\bibitem[{Jha et~al.(2015)Jha, Coke, and Radev}]{Surveyor}
Rahul Jha, Reed Coke, and Dragomir Radev. 2015.
\newblock Surveyor: A system for generating coherent survey articles for
  scientific topics.
\newblock In \emph{Proceedings of the Twenty-Ninth AAAI}, AAAI'15, pages
  2167--2173.

\bibitem[{Jie et~al.(2018)Jie, Chengzhi, Mengying, and Sanhong}]{CitationAS}
Wang Jie, Zhang Chengzhi, Zhang Mengying, and Deng Sanhong. 2018.
\newblock Citationas: A tool of automatic survey generation based on citation
  content*.
\newblock \emph{Journal of Data and Information Science}, 3(2).

\bibitem[{Luan et~al.(2018)Luan, He, Ostendorf, and Hajishirzi}]{Luan2018}
Yi~Luan, Luheng He, Mari Ostendorf, and Hannaneh Hajishirzi. 2018.
\newblock Multi-task identification of entities, relations, and coreference for
  scientific knowledge graph construction.
\newblock In \emph{Proceedings of EMNLP 2018}, pages 3219--3232.

\bibitem[{Paice(1981)}]{Paice:1980:AGL:636669.636680}
C.~D. Paice. 1981.
\newblock The automatic generation of literature abstracts: An approach based
  on the identification of self-indicating phrases.
\newblock In \emph{Proceedings of the 3rd Annual ACM SIGIR}, SIGIR '80, pages
  172--191.

\bibitem[{Paice and Jones(1993)}]{Paice:1993:IIC:160688.160696}
Chris~D. Paice and Paul~A. Jones. 1993.
\newblock \href {https://doi.org/10.1145/160688.160696} {The identification of
  important concepts in highly structured technical papers}.
\newblock In \emph{Proceedings of the 16th Annual International ACM SIGIR},
  SIGIR '93, pages 69--78, New York, NY, USA. ACM.

\bibitem[{Paik and Oard(2014)}]{Paik:2014:FMW:2661829.2661957}
Jiaul~H. Paik and Douglas~W. Oard. 2014.
\newblock A fixed-point method for weighting terms in verbose informational
  queries.
\newblock CIKM '14, pages 131--140, New York, NY, USA. ACM.

\bibitem[{Qazvinian and Radev(2008)}]{Qazvinian:2008:SPS:1599081.1599168}
Vahed Qazvinian and Dragomir~R. Radev. 2008.
\newblock \href {http://dl.acm.org/citation.cfm?id=1599081.1599168} {Scientific
  paper summarization using citation summary networks}.
\newblock In \emph{Proceedings of the 22Nd International Conference on
  Computational Linguistics - Volume 1}, COLING '08, pages 689--696.

\bibitem[{Rubinstein and Kroese(2004)}]{Rubinstein:2004:CEM:1014902}
Reuven~Y. Rubinstein and Dirk~P. Kroese. 2004.
\newblock \emph{The Cross Entropy Method: A Unified Approach To Combinatorial
  Optimization, Monte-carlo Simulation}.
\newblock Springer-Verlag, Berlin, Heidelberg.

\bibitem[{Singh et~al.(2018)Singh, Dogga, Patro, Barnwal, Dutt, Haldar, Goyal,
  and Mukherjee}]{CLscholar}
Mayank Singh, Pradeep Dogga, Sohan Patro, Dhiraj Barnwal, Ritam Dutt, Rajarshi
  Haldar, Pawan Goyal, and Animesh Mukherjee. 2018.
\newblock Cl scholar: The acl anthology knowledge graph miner.
\newblock In \emph{Proceedings of the NAACL 2018}.

\bibitem[{Xu et~al.(2009)Xu, Jones, and Wang}]{Xu:2009:QDP:1571941.1571954}
Yang Xu, Gareth~J.F. Jones, and Bin Wang. 2009.
\newblock Query dependent pseudo-relevance feedback based on wikipedia.
\newblock In \emph{Proceedings of the 32Nd International ACM SIGIR}, pages
  59--66.

\bibitem[{Yasunaga et~al.(2019)Yasunaga, Kasai, Zhang, Fabbri, Li, Friedman,
  and Radev}]{Yasunaga2019ScisummNetAL}
Michihiro Yasunaga, Jungo Kasai, Rui Zhang, Alexander~Richard Fabbri, Irene Li,
  Dan Friedman, and Dragomir~R. Radev. 2019.
\newblock Scisummnet: A large annotated corpus and content-impact models for
  scientific paper summarization with citation networks.
\newblock In \emph{AAAI 2019}.

\end{thebibliography}
\bibliographystyle{acl_natbib}

\end{document}